\documentclass{article}
\usepackage{arxiv}

\usepackage[utf8]{inputenc} 
\usepackage[T1]{fontenc}    
\usepackage{hyperref}       
\usepackage{url}            
\usepackage{booktabs}       
\usepackage{amsfonts}       
\usepackage{nicefrac}       
\usepackage{microtype}      
\usepackage{lipsum}		
\usepackage{graphicx}
\usepackage{natbib}
\usepackage{doi}

\usepackage{multirow,booktabs,graphicx}
\usepackage{algorithm,algorithmic}
\usepackage{tikz}
\usepackage{amsmath,amssymb,amsthm,mathtools}
\newtheorem{theorem}{Theorem}
\newtheorem{proposition}{Proposition}

\title{Physically Admissible Generative Flow Matching for Sparse Flow Reconstruction}  

\author{
 Yuhao Wang \\
  National University of Defense Technology\\
  Changsha, China\\
  \texttt{wangyuhao24@nudt.edu.cn} \\
   \And
 Yuanxi Peng \\
  National University of Defense Technology\\
  Changsha, China\\
  \texttt{pyx@nudt.edu.cn} \\
}

\date{}

\hypersetup{
pdftitle={Physically Admissible Generative Flow Matching for Sparse Flow Reconstruction},
pdfsubject={q-bio.NC, q-bio.QM},
pdfauthor={Yuhao Wang, Yuanxi Peng},
pdfkeywords={First keyword, Second keyword, More},
}

\begin{document}
\maketitle

\begin{abstract}
    Reconstructing full fluid velocity fields from sparse, localized observations is a severely ill-posed inverse problem: most of the spatial domain remains unobserved, discrete conservation laws must hold strictly, and predictive uncertainty must be reliable.
    Existing generative models typically enforce physical constraints only at the terminal output or via soft penalties, leaving intermediate generative trajectories unconstrained.
    We introduce Feasible Subspace Consistency (FSC), a framework that preserves physical admissibility as an invariant of continuous generative dynamics: tangent-space matching confines the entire ODE trajectory to the discrete conservation manifold, while mass-orthonormal coordinates preserve the kinetic-energy metric.
    Sparse reconstruction then reduces to amortized posterior transport in the latent space of a frozen FSC prior, ensuring every posterior sample inherits incompressibility and boundary conditions without per-sample projection loops.
    We validate FSC on a periodic Kolmogorov flow, where exact spectral projection isolates path tangency and metric isometry, and a wall-bounded cylinder wake with immersed obstacles, realized via matrix-free iterative projection.
    FSC prevents the mode-coverage collapse induced by terminal projection (increasing mode coverage from $0.000$ to $0.790$) while eliminating high-wavenumber spectral distortion.
    In sparse reconstruction, amortized FSC reduces global relative error by $50\%$ and $46\%$, and sensor residuals by $63\%$ and $55\%$, relative to the unconditioned prior, while delivering calibrated, spatially heterogeneous uncertainty.
\end{abstract}


\section{Introduction}
\label{sec:intro}

    Reconstructing full flow fields from sparse, localized sensor measurements is a foundational inverse problem in fluid mechanics. In practical systems, physical and economic constraints typically limit instrumentation to a handful of point probes or pressure taps covering less than $1\%$ of the spatial domain~\cite{mo2024reconstructing}. Such severe sparsity renders full-field reconstruction fundamentally ill-posed.

    Formally, given noisy sensor measurements $y = \mathcal{H} u + \epsilon \in \mathbb{R}^M$ under operating conditions $c$ (e.g., Reynolds number), our goal is to characterize the posterior $p(u \mid y, c)$ over the full-domain velocity field $u \in \mathbb{R}^D$ ($M \ll D$), such that every candidate realization satisfies the discrete conservation laws. Solving this physics-constrained Bayesian inverse problem presents two core challenges. First, because $M \ll D$, an extensive null space allows infinitely many states to explain the same observations, while unobserved modes carry turbulent fluctuations that deterministic point estimates average out. Second, physical admissibility is a hard constraint: incompressibility, along with wall and solid boundary conditions, must hold up to solver tolerance.

    Existing approaches address these challenges through structural compromises. Soft penalties reduce equation residuals during training but cannot guarantee discrete constraint satisfaction~\cite{raissi2019physics,shu2023physics,guo2024enhancing,bastek2025physics,baldan2025pbfm}. Hard-constraint generative models are largely restricted to simply connected or idealized 2D domains~\cite{mucke2025physics,genuist2026divergence,li2026project}, leaving wall-bounded geometries with immersed solids and amortized sparse reconstruction unaddressed. Meanwhile, POD-based estimators and EnKF assume linear or Gaussian structure~\cite{taira2017modal,evensen1994sequential}; Bayesian methods remain costly~\cite{stuart2010inverse}; and test-time guidance pulls states off the feasible manifold~\cite{parikh2026dflow,song2026flow}. Crucially, even when hard constraints are enforced, they are typically applied only at the terminal output. Yet endpoint feasibility does not imply trajectory feasibility: terminal projection can yield an admissible final sample even after the underlying generative distribution has collapsed or distorted its kinetic-energy spectrum.

    In this work, we argue that physical feasibility should be preserved as an invariant of the generative dynamics rather than imposed as a post-processing operation. 
    We introduce Feasible Subspace Consistency (FSC), a principled framework reconciling continuous generative flows with discrete physical conservation laws. 
    Under FSC, initial states are drawn on the discrete physical manifold, the generative velocity field lies strictly within the constraint tangent space, and metric isometry is preserved via mass-orthonormal coordinates. 
    Consequently, all intermediate states, terminal samples, and the generative distribution remain confined to the feasible manifold. Building on this invariant, we freeze the learned prior and condition on observations exclusively in its latent space, ensuring every posterior sample inherits the prior's physical feasibility. 

    We evaluate FSC on a periodic Kolmogorov flow and a wall-bounded cylinder wake. 
    Relative to the unconditioned prior, amortized FSC reduces global reconstruction error by roughly $50\%$ and $46\%$ and sensor residuals by $63\%$ and $55\%$, while preventing the mode-coverage collapse induced by endpoint projection in unconstrained models ($0.000 \to 0.790$). Pairing this amortized posterior with lightweight latent refinement (FSC + APLR) further achieves state-of-the-art accuracy among physically feasible methods across both linear and nonlinear observation regimes.

    We make three main contributions:
    (1) We propose Feasible Subspace Consistency (FSC), proving that continuous tangent-space matching guarantees feasibility along the entire ODE trajectory while mass-orthonormal coordinates preserve the kinetic-energy metric to prevent high-wavenumber spectral distortion.
    (2) We formulate sparse flow reconstruction as latent posterior transport over a frozen physical prior, proving that the pushforward posterior inherits exact physical feasibility while keeping sensor conditioning fully amortized without per-sample test-time optimization.
    (3) We establish theoretical guarantees showing that posterior variance contracts along observable directions while being preserved in sensor blind spots, and demonstrate near-nominal calibration with spread retained in unobserved regions.

\section{Related Work}
\label{sec:related_work}

\paragraph{Enforcing physical constraints in flow-field modeling.}
    Integrating conservation laws into deep generative models spans soft residual regularization and hard kinematic constraints. Physics-informed neural networks and diffusion models~\cite{raissi2019physics,shu2023physics,guo2024enhancing} penalize equation residuals in the training loss, which introduces multi-objective trade-offs and cannot satisfy discrete conservation laws up to solver tolerance. Hard-constraint architectures embed incompressibility via projection layers or stream functions~\cite{mohan2023embedding}, though stream functions are confined to simply connected 2D domains without internal obstacles. Recent divergence-free diffusion and operator models achieve hard feasibility on idealized 2D domains~\cite{mucke2025physics,genuist2026divergence,li2026project}, yet omit wall-bounded geometries with immersed solids and amortized sparse reconstruction. Meanwhile, flow-matching and continuous normalizing flow models~\cite{lipman2023flow,liu2023flow,chen2018neural} demonstrate strong expressive power for turbulent kinematics~\cite{parikh2026conditional,li2026generative}, but two gaps remain: enforcing constraints only at the terminal output allows intermediate ODE states to drift off the manifold, and parameterizing flows without alignment to the physical kinetic-energy metric distorts spectral energy across scales.

\paragraph{Sparse flow-field reconstruction and data assimilation.}
    Classical data assimilation methods, such as POD with linear stochastic estimation (POD-LSE)~\cite{taira2017modal} and ensemble Kalman filtering (EnKF)~\cite{evensen1994sequential}, offer analytical tractability but rely on linear-Gaussian assumptions that struggle with strongly nonlinear, multi-modal vortex dynamics. Deep surrogates~\cite{dupuy2023modeling} and neural operators~\cite{li2021fourier,kovachki2023neural} provide fast deterministic inference, yet regression targets the conditional mean. In the severely underdetermined regime, averaging over the extensive null space attenuates localized turbulent fluctuations and omits predictive uncertainty. Recent generative approaches recover fine-scale structures from sparse measurements~\cite{oommen2025learning,amoros2026guiding}, but typically require expensive per-sample guidance or tightly couple the model to a fixed sensor layout.

\paragraph{Bayesian posterior inference with generative priors.}
    Generative models increasingly serve as data-driven priors for Bayesian inverse problems. One paradigm steers an unconditional trajectory using observation-likelihood gradients, such as diffusion posterior sampling (DPS)~\cite{chung2023dps}, source-space SGLD~\cite{parikh2026dflow}, and flow-matching posterior sampling~\cite{song2026flow}. However, these methods backpropagate through the numerical ODE (or adjoint PDE) at every sampling step, and repeated unconstrained updates can push intermediate states off the feasible manifold defined by walls and obstacles. A second paradigm trains conditional generative models end-to-end on paired observation-state data~\cite{parikh2026conditional}, coupling the prior to a specific observation operator such that reconfiguring sensor placements or noise levels often degrades accuracy or necessitates retraining.

\paragraph{Positioning of FSC.}
    Unlike soft-constrained~\cite{raissi2019physics,shu2023physics,guo2024enhancing} or terminal-projection~\cite{mucke2025physics} methods, FSC guarantees trajectory-wide physical feasibility on general wall-bounded domains by construction. Unlike test-time guidance~\cite{parikh2026dflow,song2026flow} or sensor-coupled conditional training~\cite{parikh2026conditional}, FSC freezes the admissible prior and performs amortized posterior transport in its latent space, avoiding both per-sample ODE backpropagation and sensor-layout retraining.

\section{Method}
\label{sec:method}

    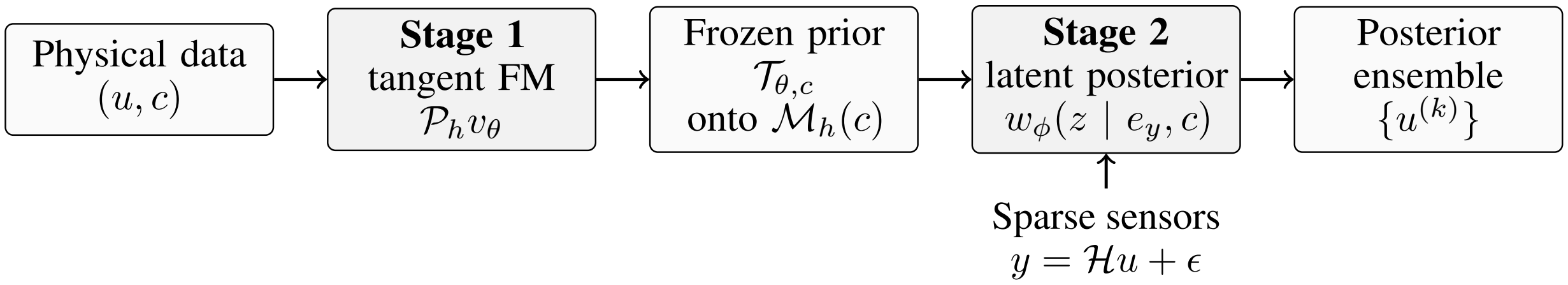
\begin{figure*}[t]
    \centering
    \resizebox{0.98\linewidth}{!}{%
    \begin{tikzpicture}[
        box/.style={draw, rounded corners=2pt, align=center, minimum height=10mm, inner sep=3pt},
        stage/.style={box, fill=black!5, text width=22mm},
        flow/.style={box, fill=black!2, text width=22mm},
        ar/.style={->, line width=0.8pt}
    ]
        \node[flow]  (data)  at (0.0,0) {Physical data\\$(u,c)$};
        \node[stage] (s1)    at (2.9,0) {\textbf{Stage 1}\\tangent FM\\$\mathcal{P}_h v_\theta$};
        \node[flow]  (prior) at (5.8,0) {Frozen prior\\$\mathcal{T}_{\theta,c}$\\onto $\mathcal{M}_h(c)$};
        \node[stage] (s2)    at (8.7,0) {\textbf{Stage 2}\\latent posterior\\$w_\phi(z\mid e_y,c)$};
        \node[flow]  (out)   at (11.6,0) {Posterior\\ensemble\\$\{u^{(k)}\}$};
        \node[align=center] (sensor) at (8.7,-1.45) {Sparse sensors\\$y=\mathcal{H}u+\epsilon$};

        \draw[ar] (data) -- (s1);
        \draw[ar] (s1) -- (prior);
        \draw[ar] (prior) -- (s2);
        \draw[ar] (s2) -- (out);
        \draw[ar] (sensor) -- (s2);
    \end{tikzpicture}}
    \caption{Overview of FSC. Stage~1 learns a tangent-space flow and defines a frozen decoder whose image lies on $\mathcal{M}_h(c)$. Stage~2 maps sparse observations to the prior latent space; every decoded posterior sample inherits the physical constraints under an exact projector (and up to solver tolerance in numerical realization).}
    \label{fig:method_overview}
    \end{figure*}

\subsection{Problem Formulation}
\label{sec:method_setup}

    Let $\Omega \subset \mathbb{R}^2$ be a fluid domain discretized onto an $N_x \times N_y$ grid containing solid obstacles $\Omega_s$ and walls $\partial\Omega_w$, and let $u \in \mathbb{R}^D$ ($D = 2 N_x N_y$) denote a discrete velocity field conditioned on operating conditions $c$ (e.g., the Reynolds number $\mathrm{Re}$).

    Given a discrete conservation and boundary operator $\mathcal{C}_h(c) \in \mathbb{R}^{K_c \times D}$ and an affine target vector $d_h(c) \in \mathbb{R}^{K_c}$, the physically admissible state space forms an affine subspace:
    \begin{equation}
    \mathcal{M}_h(c) \coloneqq \left\{ u \in \mathbb{R}^D \;\middle|\; \mathcal{C}_h(c) u = d_h(c) \right\},
    \label{eq:manifold_def}
    \end{equation}
    where $\mathcal{C}_h(c) u = d_h(c)$ enforces:
    (1) discrete solenoidal incompressibility $\mathcal{D}_h u = 0$ over fluid cells;
    (2) no-slip and impermeability conditions on physical walls $\mathcal{B}_{\mathrm{wall}} u = 0$;
    (3) zero velocity inside and on immersed solids $\mathcal{B}_{\mathrm{solid}} u = 0$; and
    (4) parameter-dependent Dirichlet inflows $\mathcal{B}_{\mathrm{in}} u = u_{\mathrm{in}}(c)$.
    At every $u \in \mathcal{M}_h(c)$, the tangent space coincides with the null space of the homogeneous constraint operator, $\mathcal{T}_{u}\mathcal{M}_h(c) = \ker(\mathcal{C}_h(c))$.

    Let $M_h \in \mathbb{R}^{D \times D}$ denote the symmetric positive-definite discrete mass metric tensor induced by fluid-cell volumes. We assume $\mathcal{C}_h(c)$ has full row rank and $\mathcal{M}_h(c)$ is nonempty. The canonical $M_h$-orthogonal projector onto $\ker(\mathcal{C}_h(c))$ is the idempotent linear operator $\mathcal{P}_h \colon \mathbb{R}^D \to \ker(\mathcal{C}_h(c))$ defined by $\mathcal{P}_h v \coloneqq \arg\min_{w \in \ker(\mathcal{C}_h)} \|v - w\|_{M_h}^2$, which admits the closed form
    \begin{equation}
        \mathcal{P}_h v = v - M_h^{-1}\mathcal{C}_h^\top
           \left( \mathcal{C}_h M_h^{-1} \mathcal{C}_h^\top \right)^{-1} \mathcal{C}_h v.
    \label{eq:projector_def}
    \end{equation}
    By construction, $\mathcal{C}_h \mathcal{P}_h = 0$, $\mathcal{P}_h^2 = \mathcal{P}_h$, and $\mathcal{P}_h v = v$ for all $v \in \ker(\mathcal{C}_h)$.
    For notational brevity, we suppress the dependence on $c$ in $\mathcal{C}_h$ and $\mathcal{P}_h$ when clear from context; in numerical implementations, the iterative realization $\mathcal{P}_h^{\delta}$ satisfies the constraints up to solver tolerance $\delta$.

    A sparse sensor layout induces a measurement operator $\mathcal{H}\colon\mathbb{R}^D\to\mathbb{R}^M$ with $M\ll D$, yielding noisy observations $y=\mathcal{H}u+\epsilon$ with $\epsilon\sim\mathcal{N}(0,\Sigma_\epsilon)$ and $\Sigma_\epsilon=\sigma_\epsilon^2 I_M$.
    Our objective is to characterize the posterior $p(u\mid y,c)$ such that every sample lies on $\mathcal{M}_h(c)$. Figure~\ref{fig:method_overview} summarizes the resulting two-stage framework.

\subsection{Stage 1: Path-Feasible Prior}
\label{sec:method_prior}

    Flow-matching models transport a base distribution to data along linear interpolants $x_t = (1-t)x_0 + t x_1$. Conventional physics-constrained variants typically learn an unconstrained velocity field and project only the terminal state $\hat{u}=\mathcal{P}_h x_1$.
    The following theorem formalizes why pathwise tangency, rather than endpoint projection alone, confines the entire generative trajectory to $\mathcal{M}_h(c)$.

    \begin{theorem}[Pathwise Feasibility]
    \label{thm:path_feasibility}
    Suppose $v_\theta(\cdot, t, c)$ is continuous in $t$ and globally Lipschitz continuous in $u$, and consider the projected neural ODE
    \begin{equation}
        \frac{d u_t}{d t} = \mathcal{P}_h\, v_\theta(u_t,t,c), \quad u_0 \sim p_0(u_0),
        \label{eq:prior_ode}
    \end{equation}
    where $\operatorname{supp}(p_0) \subseteq \mathcal{M}_h(c)$.
    Under exact $\mathcal{P}_h$, the unique solution satisfies $u_t \in \mathcal{M}_h(c)$ for all $t \in [0,1]$, and the induced terminal distribution satisfies $q_\theta(\mathcal{M}_h(c)\mid c)=1$.
    \end{theorem}

    \begin{proposition}[Isometric Transport]
    \label{prop:isometry}
    Let $d\coloneqq\dim\ker(\mathcal{C}_h(c))$, let $Q_h(c) \in \mathbb{R}^{D \times d}$ be an $M_h$-orthonormal basis of $\ker(\mathcal{C}_h(c))$ satisfying $Q_h^\top M_h Q_h=I_d$, and let $\ell_h(c)\in\mathcal{M}_h(c)$ be any particular solution.
    Then the affine decoder
    \begin{equation}
    F_c(z) \coloneqq \ell_h(c)+Q_h(c)z
    \end{equation}
    is a metric isometry between $(\mathbb{R}^d, \|\cdot\|_2)$ and $(\mathcal{M}_h(c), d_{M_h})$, where $d_{M_h}(u,v)\coloneqq\sqrt{(u-v)^\top M_h(u-v)}$ is the kinetic-energy metric; that is, for all $z_a,z_b\in\mathbb{R}^d$,
    \begin{equation}
    d_{M_h}\big(F_c(z_a),F_c(z_b)\big)=\|z_a-z_b\|_2.
    \end{equation}
    Consequently, for any two probability measures $\mu,\nu$ on $\mathbb{R}^d$ with finite $p$-th moments ($p \ge 1$),
    \begin{equation}
    \mathcal{W}_{p,M_h}\big(F_{c\#}\mu,F_{c\#}\nu\big)=\mathcal{W}_p(\mu,\nu).
    \label{eq:wasserstein_isometry}
    \end{equation}
    \end{proposition}

    \paragraph{Tangent-space flow matching.}
    We construct a feasible base distribution by projecting standard Gaussian noise onto $\mathcal{M}_h(c)$, i.e., $x_0 = \ell_h(c) + \mathcal{P}_h\xi$ with $\xi\sim\mathcal{N}(0,I_D)$, ensuring $\operatorname{supp}(p_0)\subseteq\mathcal{M}_h(c)$, and draw target states $x_1\sim p_{\mathrm{data}}(\cdot\mid c) \subset \mathcal{M}_h(c)$.
    Because $\mathcal{M}_h(c)$ is affine, the linear interpolant $x_t = (1-t)x_0 + t x_1$ remains in $\mathcal{M}_h(c)$ for all $t \in [0,1]$, and its velocity $\dot x_t = x_1-x_0$ lies in $\ker(\mathcal{C}_h)$.
    Writing $v_{\theta,h}(u,t,c)\coloneqq\mathcal{P}_h v_\theta(u,t,c)$ so that $\mathcal{C}_h v_{\theta,h}=0$, we train $v_{\theta,h}$ via the FSC tangent-space flow-matching objective:
    \begin{equation}
    \mathcal{L}_{\mathrm{prior}}(\theta)
    = \mathbb{E}_{t, x_0, x_1}
    \left\|
    \mathcal{P}_h v_\theta(x_t,t,c)-\mathcal{P}_h(\dot x_t)
    \right\|_{M_h}^2 .
    \label{eq:loss_prior}
    \end{equation}
    
    Since $\dot x_t\in\ker(\mathcal{C}_h)$, we have $\mathcal{P}_h(\dot x_t)=\dot x_t$, so the projector effectively acts only on the network prediction.
    On periodic domains, this objective is evaluated directly in latent coordinates by substituting $x_t=F_c(z_t)$; by Proposition~\ref{prop:isometry}, the $M_h$-norm reduces identically to the Euclidean norm, enabling training in the physical kinetic-energy geometry without coordinate rescaling or high-wavenumber spectral distortion.

\subsection{Stage 2: Amortized Posterior Transport}
\label{sec:method_posterior}

    Once trained, the frozen FSC prior induces a deterministic, invertible decoder $\mathcal{T}_{\theta,c}\colon\mathcal{Z}\to\mathcal{M}_h(c)$ via $u=\mathcal{T}_{\theta,c}(z)$, defined by integrating Eq.~\eqref{eq:prior_ode} from $t=0$ to $t=1$. Under the Lipschitz condition of Theorem~\ref{thm:path_feasibility}, $\mathcal{T}_{\theta,c}$ is a homeomorphism onto $\mathcal{M}_h(c)$.
    The parameterization of $\mathcal{Z}$ depends on domain topology:
    on periodic domains, $\mathcal{Z}=\mathbb{R}^d$ corresponds to the explicit mass-orthonormal chart of Proposition~\ref{prop:isometry};
    on wall-bounded domains where an explicit basis is unavailable, we set $\mathcal{Z}=\mathcal{M}_h(c) \subset \mathbb{R}^D$ directly in feasible physical space and evaluate $\mathcal{T}_{\theta,c}^{-1}$ via reverse-time integration of Eq.~\eqref{eq:prior_ode}. In this implicit-chart setting, feasibility holds up to the iterative tolerance $\mathcal{P}_h^{\delta}$, with a finite-tolerance error bound derived in the supplementary material.
    Crucially, $\mathcal{T}_{\theta,c}$ remains frozen as a reusable physical decoder: observation conditioning adapts solely the latent distribution without retraining or altering the prior.

    \begin{theorem}[Pushforward Equivalence and Feasibility]
    \label{thm:posterior_feasibility}
    Assume $\mathcal{T}_{\theta,c}\colon\mathcal{Z}\to\mathcal{M}_h(c)$ is a bijection, let the generative physical prior be the pushforward $p_\theta(u\mid c)=\mathcal{T}_{\theta,c\#}\phi$ of a base density $\phi(z)$ on $\mathcal{Z}$, and define the latent conditional posterior
    \begin{equation}
    \pi_\theta(z\mid y,c)
    \propto p\big(y\;\big|\;\mathcal{T}_{\theta,c}(z)\big)\phi(z).
    \label{eq:latent_posterior}
    \end{equation}
    Under exact $\mathcal{P}_h$, the pushforward of $\pi_\theta(\cdot\mid y,c)$ through $\mathcal{T}_{\theta,c}$ coincides with the Bayes posterior induced by $p_\theta$:
    \begin{equation}
    \mathcal{T}_{\theta,c\#}\pi_\theta(\cdot\mid y,c)=p_\theta(u\mid y,c),
    \label{eq:pushforward_equiv}
    \end{equation}
    and every posterior realization satisfies the physical constraints almost surely:
    \begin{equation}
    p_\theta\big(\mathcal{M}_h(c)\;\big|\;y,c\big)=1.
    \label{eq:posterior_support}
    \end{equation}
    \end{theorem}

    \begin{proposition}[Local Contraction and Blind-Spot Retention]
    \label{prop:uncertainty_contraction}
    Assume a local Gaussian latent prior $z\sim\mathcal{N}(z_0,\Sigma_z)$ with $\Sigma_z\succ0$, let $J_\theta\coloneqq\nabla_z\mathcal{T}_{\theta,c}(z_0)$ be the decoder Jacobian, and linearize the forward model around $z_0$ as $y\approx\mathcal{H}\mathcal{T}_{\theta,c}(z_0)+A(z-z_0)+\epsilon$ with $A\coloneqq\mathcal{H}J_\theta$.
    Then observation information contracts posterior variance along $\operatorname{range}(A^\top)$, whereas for every $w\in\ker(A\Sigma_z)$ (which coincides with $\ker(A)$ when $\Sigma_z=\sigma_0^2 I$), the posterior variance is preserved:
    \begin{equation}
    w^\top\Sigma_{z\mid y}w=w^\top\Sigma_z w.
    \label{eq:blind_spot_variance}
    \end{equation}
    \end{proposition}
    Under local linearization, Proposition~\ref{prop:uncertainty_contraction} highlights a fundamental contrast: deterministic point estimators collapse variance across all directions—including sensor blind spots $\ker(A)$—thereby over-smoothing unresolved turbulent scales, whereas the FSC posterior ensemble preserves prior uncertainty along unobserved modes.
    Bounds on prior-to-posterior error propagation in total variation and $\mathcal{W}_{1,M_h}$ are provided in the supplementary material.

    \paragraph{Amortized latent flow matching.}
    Rather than running test-time MCMC or solving per-instance latent optimizations, we train an amortized conditional velocity network $w_\phi\colon\mathcal{Z}\times[0,1]\times\mathbb{R}^{C_e}\to T\mathcal{Z}$, where $T\mathcal{Z}=\mathbb{R}^d$ on periodic domains and $T\mathcal{Z}=\ker(\mathcal{C}_h(c))$ on wall-bounded domains.
    We invert the frozen ODE to obtain latent targets $z=\mathcal{T}_{\theta,c}^{-1}(u)$ and encode measurements via $e_y=\operatorname{Enc}(y,\mathcal{H})\in\mathbb{R}^{C_e}$. Drawing base noise $\xi\sim p_0$ (with $p_0=\mathcal{N}(0,I_d)$ on periodic domains and $p_0=\ell_h(c)+\mathcal{P}_h\mathcal{N}(0,I_D)$ on wall-bounded domains), we form $z_t\coloneqq(1-t)\xi+t z$ with target velocity $v^\star\coloneqq z-\xi \in T\mathcal{Z}$, and minimize
    \begin{equation}
    \mathcal{L}_{\mathrm{posterior}}(\phi)=\mathbb{E}_{t,\xi,z}\big\|w_\phi(z_t,t,c,e_y)-v^\star\big\|_2^2.
    \label{eq:loss_post}
    \end{equation}
    Because conditioning occurs entirely in $\mathcal{Z}$ while decoding passes through the frozen feasible map $\mathcal{T}_{\theta,c}$, physical feasibility holds by construction regardless of posterior amortization error.

\section{Experiments}
\subsection{Experimental Setup}
\label{sec:exp_setup}

\paragraph{Physical Benchmark Systems and Splits.}

    We evaluate FSC across two distinct flow topologies:
    (1) System~1 (Canonical Periodic Turbulence): Two-dimensional Kolmogorov flow governed by the incompressible Navier--Stokes equations on a flat torus $\mathbb{T}^2 = [0, 2\pi]^2$, discretized on a $128 \times 128$ uniform grid ($D = 32{,}768$), driven by monochromatic body forcing ($F=1$, $k_f=4$) with linear drag $\alpha=0.1$ and viscosity $\nu = 1/\mathrm{Re}$; the dataset spans $45$ Reynolds-number groups ($\mathrm{Re} \in [60, 650]$, $32{,}500$ frames).
    (2) System~2 (Wall-Bounded Flow Past a Cylinder): Viscous flow past an immersed circular cylinder ($D_{\mathrm{cyl}}=1.0$) in a channel $\Omega = [0, 20.0] \times [0, 10.0]$, discretized on a $256 \times 128$ staggered MAC grid ($D = 65{,}536$), with rigid no-slip walls, body-force driving $G(\mathrm{Re})$, and Brinkman obstacle penalization; the dataset spans $42$ trajectories ($\mathrm{Re} \in [70, 130]$, $16{,}842$ frames).
    For System~1, training, validation, and test splits are drawn from $\mathrm{Re}\in[120,500]$ ($20$ groups) with an $8{:}1{:}1$ split within each group; System~2 uses $18$ training, $7$ validation, and $6$ in-distribution test trajectories at $\mathrm{Re}\in\{80,100,120\}$. Per-split frame counts are detailed in the supplementary material.

\paragraph{Reconstruction Protocol.}
    Standard sparse reconstructions use $S=64$ uniformly distributed sensors measuring both velocity components ($M=2S=128$), with additive white Gaussian noise $\sigma_\epsilon=0.01$ and $K=32$ posterior samples. System~1 evaluates $20$ in-distribution held-out cases (test-in), while the indirect/nonlinear-operator experiments evaluate $16$ interpolation cases at unseen odd Reynolds numbers ($130$--$430$). System~2 evaluates $16$ in-distribution test cases from unseen random seeds. All reconstructions use the frozen Stage~1 prior. During Stage~2 posterior training, sensor counts are randomized over $S\in[8,128]$ and noise levels over $\sigma_\epsilon\in\{0,0.01,0.05,0.1\}$ on System~1, and over $S\in[16,64]$ and $\sigma_\epsilon\in\{0,0.01,0.05\}$ on System~2; a single amortized network is evaluated across unseen sensor counts and noise levels without retraining. Sensor-count sensitivity analyses are reported in the supplementary material.

\paragraph{Baselines and Optional Refinement.}
    We compare against two primary baselines:
    (1) Supervised regression: a deterministic neural operator/network trained end-to-end on paired observations and full fields; and
    (2) EnKF+proj: a validation-tuned ensemble Kalman filter ($K=32$) operating on the frozen prior, followed by a terminal feasible projection.
    In addition, we evaluate an optional test-time refinement variant, FSC + APLR (Amortized Posterior with Latent Refinement), which performs $20$ gradient steps on the observation likelihood in the latent space $\mathcal{Z}$ initialized from the amortized FSC posterior; because optimization occurs strictly in $\mathcal{Z}$ ahead of the frozen decoder $\mathcal{T}_{\theta,c}$, physical feasibility is preserved throughout refinement.

\paragraph{Evaluation Metrics.}
    We report four families of metrics:
    (1) Physical feasibility: relative divergence $\delta_{\mathrm{div}}(u)=\|\mathcal{D}_h u\|_{\mathrm{rms}}/\|u\|_{\mathrm{rms}}$;
    (2) Reconstruction accuracy: relative $L_2$ error $e_2(u)=\|u-\bar{u}\|_2/\|u\|_2$ and relative sensor residual $e_y(u)=\|\mathcal{H}\bar{u}-y\|_2/\|y\|_2$, where $\bar{u}$ denotes the posterior ensemble mean;
    (3) Distributional fidelity: sliced Wasserstein distance (SWD), outermost shell energy ratio $k_{13}$ (ideal $1.0$), out-of-band energy fraction (OOB), mode coverage (fraction of data-occupied kinetic-energy bins populated by generated samples), and energy coverage (E-cov, fraction of reference kinetic energy captured); and
    (4) Uncertainty and calibration: continuous ranked probability score (CRPS), empirical $90\%$ coverage $\mathrm{cov90}$ (fraction of grid coordinates falling within the predicted $90\%$ credible interval; ideal $0.900$), and spatial standard-deviation maps $s(x)$. On System~1, the feasible subspace is band-limited to $|k|\le13$, retaining $99.9\%$ of the reference kinetic energy, with $k_{13}$ denoting its outermost spectral shell.

\paragraph{Neural Backbones and Optimization.}
    System~1 uses a multiscale U-shaped Fourier neural operator (U-FNO), whose global spectral convolutions match the periodic domain, and System~2 uses a deep convolutional U-Net, whose local receptive fields accommodate walls and immersed obstacles. Both models are trained with AdamW (weight decay $10^{-4}$, $\beta_1=0.9$, $\beta_2=0.999$), a cosine learning-rate schedule from $2\times10^{-4}$ to $10^{-6}$, and exponential moving average weights (decay $0.999$). The System~1 prior trains for $80$ epochs at batch size $64$, the System~2 prior for $200$ epochs at batch size $16$, and the Stage~2 posterior transport network for $20{,}000$ steps on System~1 and $16{,}000$ on System~2 at batch size $8$. Architecture details are provided in the supplementary material.


\subsection{Physical Prior Verification and Feasibility Guarantees}
\label{sec:exp_prior}

    Before addressing sparse reconstruction, we validate the frozen Stage~1 prior across three ablation levels: where the constraint is enforced, which subspace geometry and metric are used, and whether the matrix-free realization transfers to wall-bounded wakes.

\paragraph{Constraint placement (Layer 1).}
    \begin{table}[htbp]
    \centering
    \caption{Constraint placement on System~1. {Vanilla} is an unconstrained velocity-space flow-matching model; the {+} rows add, respectively, endpoint projection of the terminal output, retraction onto the band-limited subspace ($|k|\le13$) after each ODE step, and a soft divergence penalty $\lambda\|\nabla\cdot v\|^2/\|v\|^2$. {FSC full-chain} enforces feasibility consistently across parameterization, training path, vector field, ODE state, and decoder.}
    \label{tab:constraint_placement}
    \begin{tabular}{lccc}
    \toprule
    Configuration & Mode cov. $\uparrow$ & SWD $\downarrow$ & $\delta_{\mathrm{div}}$ $\downarrow$ \\
    \midrule
    Vanilla                 & $0.790$ & $0.362$ & $3.9\times10^{-1}$ \\
    + soft penalty          & $0.799$ & $0.369$ & $9.8\times10^{-3}$ \\
    + endpoint projection   & $0.000$ & $0.116$ & $6.9\times10^{-6}$ \\
    + step retraction       & $0.317$ & $0.438$ & $7.6\times10^{-6}$ \\
    FSC full-chain          & $0.790$ & $0.073$ & $7.2\times10^{-6}$ \\
    \bottomrule
    \end{tabular}
    \end{table}

    Table~\ref{tab:constraint_placement} isolates the effect of constraint placement. Vanilla FM and Vanilla + soft penalty fail physical feasibility, exhibiting relative divergences of $3.9\times10^{-1}$ and $9.8\times10^{-3}$, respectively. Conversely, Vanilla + endpoint projection and Vanilla + step retraction achieve low divergence ($\sim 7\times10^{-6}$) but severely distort the generative distribution, collapsing mode coverage to $0.000$ and $0.317$. Only FSC full-chain satisfies both criteria simultaneously, preserving full mode coverage ($0.790$) while achieving the lowest SWD ($0.073$) and strict feasibility ($\delta_{\mathrm{div}} = 7.2\times10^{-6}$). These results confirm that physical feasibility must be maintained intrinsically along the generative trajectory rather than imposed via post-hoc projection or soft penalties.

\paragraph{Subspace geometry (Layer 2).}

    \begin{table}[htbp]
    \centering
    \caption{Subspace geometry and metric alignment on System~1. Unbounded retains full-chain feasibility without a hard spectral cutoff; band-limited $\psi$ and band-limited mass both enforce full-chain feasibility on $|k|\le13$, differing solely in the coordinate metric.}
    \label{tab:subspace_geometry}
    \begin{tabular}{lcccc}
    \toprule
    Configuration & Mode cov. $\uparrow$ & $k_{13}$ ($\to 1.0$) & OOB $\downarrow$ & $\delta_{\mathrm{div}}$ $\downarrow$ \\
    \midrule
    Unbounded             & $0.083$ & $0.887$  & $0.0351$ & $7.0\times10^{-6}$ \\
    Band-limited $\psi$   & $0.735$ & $13.719$ & $0.0000$ & $7.3\times10^{-6}$ \\
    Band-limited mass     & $0.790$ & $1.301$  & $0.0000$ & $7.2\times10^{-6}$ \\
    \bottomrule
    \end{tabular}
    \end{table}
    
    Table~\ref{tab:subspace_geometry} holds full-chain consistency fixed and varies the subspace parameterization. All three configurations achieve comparable divergence ($\delta_{\mathrm{div}}\approx 7\times10^{-6}$), but the unbounded parameterization leaks $3.51\%$ of its energy outside the resolved band (OOB $0.0351$) and suffers severe mode-coverage degradation ($0.083$). Within the band-limited subspace, stream-function ($\psi$) and mass-orthonormal coordinates differ solely in their induced metric: $\psi$-coordinates heavily amplify high-wavenumber energy, driving the band-edge shell ratio $k_{13}$ to $13.719$, whereas mass-orthonormal coordinates keep $k_{13}$ near unity ($1.301$) with zero out-of-band leakage. Mass-orthonormal parameterization thus preserves the physical kinetic-energy spectrum, directly corroborating Proposition~\ref{prop:isometry}.

\paragraph{Matrix-free realization (Layer 3).}

    \begin{table}[htbp]
    \centering
    \caption{Constraint placement on System~2 (wall-bounded cylinder wake), with rows aligned to Table~\ref{tab:constraint_placement}.}
    \label{tab:system2_constraint}
    
    \begin{tabular}{lccc}
    \toprule
    Configuration & E-cov $\uparrow$ & SWD $\downarrow$ & $\delta_{\mathrm{div}}$ $\downarrow$ \\
    \midrule
    Vanilla FM              & $0.738$ & $0.1356$ & $1.26\times10^{-1}$ \\
    + soft penalty          & $0.738$ & $0.1464$ & $2.75\times10^{-2}$ \\
    + endpoint projection   & $0.740$ & $0.1354$ & $1.55\times10^{-6}$ \\
    + step retraction       & $0.739$ & $0.1360$ & $7.05\times10^{-7}$ \\
    FSC full-chain          & $0.746$ & $0.1190$ & $6.85\times10^{-7}$ \\
    \bottomrule
    \end{tabular}
    \end{table}

    Table~\ref{tab:system2_constraint} repeats the placement ablation on the wall-bounded domain, where the matrix-free PCG projector replaces the analytic Fourier basis. Again, Vanilla FM and Vanilla + soft penalty violate incompressibility ($\delta_{\mathrm{div}} = 1.26\times10^{-1}$ and $2.75\times10^{-2}$). While Vanilla + endpoint projection and Vanilla + step retraction achieve low output divergence ($\sim 10^{-6}$), their distributional fidelity remains stagnant at the unconstrained baseline level (SWD $0.1354$ and $0.1360$ vs.\ $0.1356$). Although the performance margin is narrower than on the periodic torus due to strong boundary anchoring, FSC full-chain still achieves the lowest SWD ($0.1190$) and highest energy coverage ($0.746$).

\paragraph{Path feasibility.}

    \begin{figure}[htbp]
    \centering
    \includegraphics[width=\linewidth]{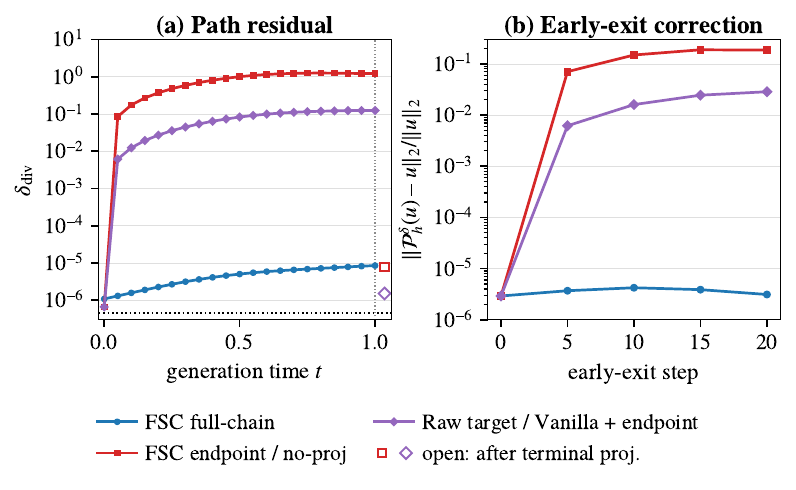}
    \caption{Inference-time trajectory ablations on System~2 ($768$ samples, $20$-step ODE solver, flux-corrected prior; FSC curves share the same checkpoint). (a) Fluid-cell relative divergence $\delta_{\mathrm{div}}$ along integration steps. (b) Relative projection correction $\|\mathcal{P}_h^\delta(u)-u\|_2/\|u\|_2$ required upon early exit at steps $0, 5, 10, 15$, and $20$.}
    \label{fig:path_residual}
    \end{figure}

    Figure~\ref{fig:path_residual} examines trajectory feasibility during ODE integration on System~2. We compare full-chain (projecting the vector field at every ODE step and at termination), endpoint-only (projecting only the final state), no projection (omitting projection entirely), and a raw-target baseline (Vanilla checkpoint with terminal projection). In panel~(a), endpoint-only and no-projection coincide along the trajectory, with divergence climbing from $6.6\times10^{-7}$ to $1.25$; the raw-target baseline drifts less ($0.12$) but remains orders of magnitude above solver tolerance, whereas full-chain maintains $\delta_{\mathrm{div}}$ within $10^{-6}$--$10^{-5}$ throughout integration. Open markers denote post-terminal projected states, and the dotted line indicates the ground-truth reference. Panel~(b) quantifies the relative projection correction $\|\mathcal{P}_h^\delta(u)-u\|_2/\|u\|_2$ at intermediate checkpoints: full-chain requires only $3$--$4\times10^{-6}$ correction at every step, compared to $0.19$ for endpoint-only and $0.029$ for raw-target. Thus, intermediate states remain physically admissible throughout generation only under full-chain consistency.

\subsection{Sparse Reconstruction and Uncertainty Quantification}
\label{sec:exp_reconstruction}

    We next evaluate Stage~2 sparse flow reconstruction by coupling observation conditioning with the frozen FSC prior, empirically verifying the pushforward equivalence and feasibility guarantees of Theorem~\ref{thm:posterior_feasibility}.

\paragraph{Reconstruction performance across both benchmarks.}

    \begin{table*}[htbp]
        \centering
        \small
        \caption{Sparse posterior reconstruction on the test-in split across both physical benchmarks ($S=64$ uniform sensors, $K=32$ ensemble members, $\sigma_\epsilon=0.01$). Unconditioned prior denotes the frozen Stage~1 generator without sensor conditioning; FSC + APLR adds $20$ latent gradient refinement steps. Boldface marks the best $e_2$ and $e_y$ among physically feasible methods; divergence entries are not ranked, as every feasible method sits at the solver floor.}
        \label{tab:reconstruction_summary}
        \begin{tabular}{lcccccc}
        \toprule
        & \multicolumn{3}{c}{\textbf{System~1 (Periodic Kolmogorov Flow)}} & \multicolumn{3}{c}{\textbf{System~2 (Wall-Bounded Cylinder Wake)}} \\
        \cmidrule(lr){2-4}\cmidrule(lr){5-7}
        Method & $e_2$ $\downarrow$ & $e_y$ $\downarrow$ & $\delta_{\mathrm{div}}$ $\downarrow$ & $e_2$ $\downarrow$ & $e_y$ $\downarrow$ & $\delta_{\mathrm{div}}$ $\downarrow$ \\
        \midrule
        Unconditioned prior          & $0.9871$ & $1.0001$ & $3.66\times10^{-6}$ & $0.3165$ & $0.2252$ & $8.33\times10^{-7}$ \\
        Supervised regression        & $0.2901$ & $0.1127$ & $5.06\times10^{-1}$          & $0.0887$ & $0.0323$ & $1.53\times10^{-1}$ \\
        EnKF+proj                    & $0.4601$ & $0.2071$ & $5.14\times10^{-6}$          & $0.1638$ & $0.0411$ & $1.18\times10^{-6}$ \\
        FSC posterior (amortized)    & $0.4968$ & $0.3730$ & $4.30\times10^{-6}$          & $0.1702$ & $0.1020$ & $1.72\times10^{-6}$ \\
        FSC + APLR (refinement)      & $\mathbf{0.2880}$ & $\mathbf{0.0850}$ & $6.10\times10^{-6}$ & $\mathbf{0.1574}$ & $\mathbf{0.0137}$ & $3.21\times10^{-6}$ \\
        \bottomrule
        \end{tabular}
    \end{table*}

\begin{figure*}[htbp]
    \centering
    \includegraphics[width=\textwidth]{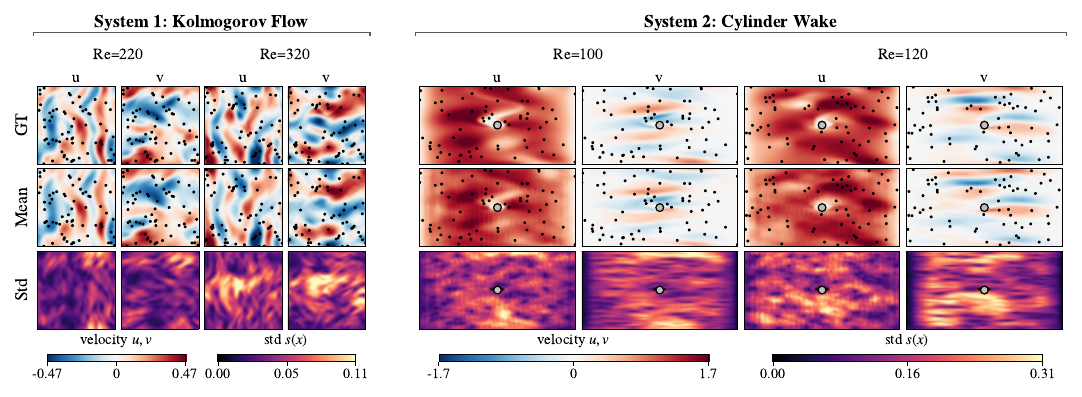}
    \caption{Qualitative sparse reconstruction with FSC + APLR on held-out test cases ($S=64$ sensors, $K=32$ posterior samples, $\sigma_\epsilon=0.01$). Left: System~1 (Kolmogorov flow) at $\mathrm{Re}=220$ and $320$ ($128^2$ grid). Right: System~2 (cylinder wake) at $\mathrm{Re}=100$ and $120$ ($256\times128$ grid). Rows: ground truth, posterior ensemble mean, and posterior standard deviation $s(x)$; columns: velocity components $u$ and $v$, with sensor locations overlaid on the GT and mean panels. Within each system, all velocity panels share one symmetric color scale and all std panels share one scale; GT and mean use identical $v_{\min}/v_{\max}$.}
    \label{fig:recon}
    \end{figure*}

    Table~\ref{tab:reconstruction_summary} summarizes test-in sparse reconstruction at $S=64$ across both physical systems. Without observation conditioning, the frozen Stage~1 prior maintains strict physical feasibility ($\delta_{\mathrm{div}} = 3.66\times10^{-6}$ and $8.33\times10^{-7}$) but exhibits high reconstruction error ($e_2 = 0.9871$ and $0.3165$). Single-pass amortized conditioning (FSC posterior) reduces $e_2$ by roughly $50\%$ on System~1 ($0.4968$) and $46\%$ on System~2 ($0.1702$), while lowering sensor residuals by $63\%$ and $55\%$ and keeping every posterior realization on $\mathcal{M}_h(c)$—achieving accuracy comparable to sequential EnKF+proj ($0.4601$ and $0.1638$) without per-step ensemble assimilation. Supervised regression attains lower point error ($0.2901$ and $0.0887$), but violates discrete conservation laws by five orders of magnitude ($\delta_{\mathrm{div}} = 5.06\times10^{-1}$ and $1.53\times10^{-1}$). Pairing the amortized FSC posterior with lightweight latent refinement (FSC + APLR) further lowers $e_2$ to $0.2880$ and $0.1574$ and $e_y$ to $0.0850$ and $0.0137$ without leaving $\mathcal{M}_h(c)$, achieving the best accuracy among all physically feasible methods on both benchmarks. Figure~\ref{fig:recon} illustrates qualitative reconstructions on held-out test cases. On System~1 ($\mathrm{Re}=220$ and $320$), the posterior mean accurately recovers dominant vortical structures, while the ensemble standard deviation $s(x)$ exhibits pronounced spatial heterogeneity aligned with unresolved shear layers. On System~2 ($\mathrm{Re}=100$ and $120$), the posterior mean cleanly reconstructs the K\'arm\'an vortex street and wake recirculation region without wall or obstacle artifacts, concentrating predictive uncertainty along the unsteady wake shear layers.

\paragraph{Uncertainty quantification.}
\label{sec:exp_uq}
    
    \begin{table}[htbp]
        \centering
        \caption{Uncertainty calibration on the test-in split. Supervised regression is deterministic, so $\mathrm{cov90}$ and CRPS are undefined.}
        \label{tab:uq_summary}
        
        \begin{tabular}{lcccc}
        \toprule
        & \multicolumn{2}{c}{System~1} & \multicolumn{2}{c}{System~2} \\
        \cmidrule(lr){2-3}\cmidrule(lr){4-5}
        Method & $\mathrm{cov90}$ $\uparrow$ & CRPS $\downarrow$ & $\mathrm{cov90}$ $\uparrow$ & CRPS $\downarrow$ \\
        \midrule
        Unconditioned prior & $0.827$ & $0.105$ & $0.776$ & $0.131$ \\
        Supervised regression & -- & -- & -- & -- \\
        EnKF+proj & $0.900$ & $0.046$ & $0.838$ & $0.065$ \\
        FSC posterior & $0.875$ & $0.051$ & $0.921$ & $0.073$ \\
        FSC + APLR (refinement) & $0.871$ & $0.028$ & $0.953$ & $0.068$ \\
        \bottomrule
        \end{tabular}
    \end{table}

\begin{figure}[htbp]
        \centering
        \includegraphics[width=\linewidth]{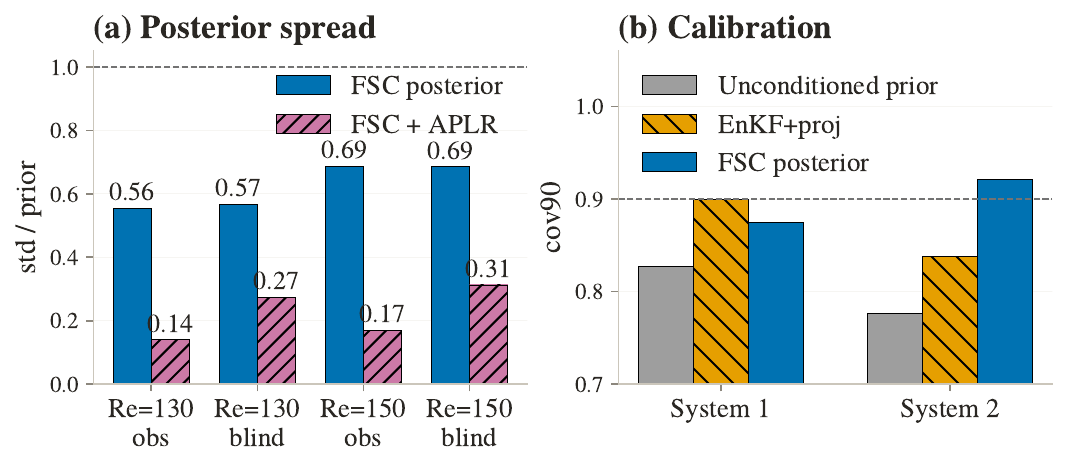}
        \caption{Uncertainty diagnostics. (a) Ratio of posterior to prior ensemble standard deviation on System~1 for two test-in cases ($\mathrm{Re}=130$ and $150$), split into observed and sensor-blind regions. (b) Empirical $90\%$ coverage ($\mathrm{cov90}$) on the test-in split for both systems; the dashed line marks the nominal $0.900$ target.}
        \label{fig:uq_main}
    \end{figure}

    Table~\ref{tab:uq_summary} summarizes calibration metrics on the test-in split. The unconditioned prior under-covers on both systems ($0.827$ and $0.776$), whereas amortized FSC conditioning brings $\mathrm{cov90}$ within $2.5$ percentage points of the nominal $0.900$ target ($0.875$ on System~1 and $0.921$ on System~2) while substantially lowering CRPS. Figure~\ref{fig:uq_main} further partitions ensemble spread into observed and sensor-blind regions. Because spatial correlations and Navier--Stokes coupling link local measurements to global modes, the amortized posterior contracts spread moderately across the domain while retaining $56$--$69\%$ of the prior standard deviation in sensor-blind regions—avoiding the variance collapse of point estimators in accordance with Proposition~\ref{prop:uncertainty_contraction}. Applying APLR further sharpens uncertainty in observed regions ($14$--$17\%$ of prior spread) while retaining $27$--$31\%$ spread in unobserved regions.

\subsection{Indirect and Nonlinear Observation Operators}
\label{sec:exp_nonlinear}

    Direct linear velocity sensing ($y=\mathcal{H}u+\epsilon$) represents a favorable regime for Kalman-type updates. To test whether latent-space conditioning and the frozen feasible decoder generalize to indirect and nonlinear measurements, we replace point velocity sensors on System~1 with two challenging observation operators: (1) pointwise velocity magnitude $y_i=\|u(x_i)\|_2+\epsilon_i$, which is nonlinear and invariant to global sign inversion ($u\to -u$), and (2) pointwise vorticity $y_i=\omega(x_i)+\epsilon_i$, a differential indirect operator. Evaluations use $S=64$, noise $\sigma_\epsilon=0.01$, and $16$ unseen Reynolds-number interpolation cases. Because the amortized network is trained solely for direct velocity sensors, FSC + APLR refines the frozen prior directly per test instance; EnKF+proj applies the corresponding forward operator ($K=32$, covariance inflation $1.5$, localization radius $0.2$); and supervised regression is retrained specifically for each operator.

    \begin{table}[htbp]
    \centering
    \caption{Indirect and nonlinear observation operators on System~1 (interpolation split). For velocity magnitude $\|u\|_2$, $e_2$, $\mathrm{cov90}$, and CRPS are evaluated after resolving global sign ambiguity. Boldface marks the best value per column, including the deterministic regression.}
    \label{tab:nonlinear}
    \begin{tabular}{llcccc}
    \toprule
    Operator & Method & $e_2$ $\downarrow$ & $e_y$ $\downarrow$ & $\mathrm{cov90}$ $\uparrow$ & CRPS $\downarrow$ \\
    \midrule
    $\|u\|_2$ (nonlinear)
    & Unconditioned prior & $0.899$ & $0.624$ & $0.895$ & $0.093$ \\
    & Supervised regression & $\mathbf{0.771}$ & $0.635$ & -- & -- \\ 
    & EnKF+proj & $0.872$ & $0.662$ & $0.966$ & $0.093$ \\
    & FSC + APLR & $0.837$ & $\mathbf{0.126}$ & $0.966$ & $\mathbf{0.084}$ \\
    \midrule
    $\omega$ (differential)
    & Unconditioned prior & $0.989$ & $4.136$ & $0.858$ & $0.102$ \\
    & Supervised regression & $\mathbf{0.591}$ & $0.918$ & -- & -- \\
    & EnKF+proj & $0.793$ & $0.861$ & $0.944$ & $0.082$ \\
    & FSC + APLR & $0.769$ & $\mathbf{0.183}$ & $0.963$ & $\mathbf{0.080}$ \\
    \bottomrule
    \end{tabular}
    \end{table}

    Table~\ref{tab:nonlinear} reports results on the interpolation split. Although task-specific supervised regression attains the lowest mean $e_2$, its deterministic predictions exhibit severe observation inconsistency ($e_y = 0.635$ and $0.918$), as the conditional mean cancels opposite-sign modes under $\|u\|_2$ and over-smooths small-scale gradients under $\omega$. EnKF+proj struggles with operator nonlinearity and non-locality, yielding observation residuals $5.3\times$ higher than FSC + APLR under $\|u\|_2$ ($0.662$ vs.\ $0.126$) and $4.7\times$ higher under $\omega$ ($0.861$ vs.\ $0.183$). In contrast, FSC + APLR achieves the lowest observation residual and best CRPS among all probabilistic methods while maintaining calibrated coverage ($0.963$--$0.966$) and strict physical feasibility—demonstrating that the frozen FSC prior transfers seamlessly to unseen observation modalities without retraining.

\section{Conclusion and Limitations}
\label{sec:conclusion}

    In this work, we introduced Feasible Subspace Consistency (FSC), a framework reconciling continuous generative flows with discrete physical conservation laws. By enforcing continuous tangent-space matching along the entire generative trajectory (Theorem~\ref{thm:path_feasibility}) and metric isometry in mass-orthonormal coordinates (Proposition~\ref{prop:isometry}), FSC confines states to $\mathcal{M}_h(c)$ up to floating-point or projector tolerance, avoiding the mode collapse and spectral distortion of post-hoc projection. Building on this invariant, sparse flow reconstruction is formulated as amortized latent posterior transport over a frozen prior (Theorem~\ref{thm:posterior_feasibility}), guaranteeing physical feasibility without observation-specific projection loops. Across Kolmogorov turbulence and cylinder wakes, amortized FSC reduces global error by roughly $50\%$ and $46\%$ and sensor residuals by $63\%$ and $55\%$ over the unconditioned prior, while delivering calibrated uncertainty, with blind-spot variance retained as Proposition~\ref{prop:uncertainty_contraction} predicts. Pairing this posterior with lightweight latent refinement (FSC + APLR) further attains the best accuracy among the compared physically feasible methods across linear and nonlinear observation regimes.

    \paragraph{Limitations.}
    First, the spectral cutoff on periodic domains caps the finest resolvable scales, while wall-bounded geometries rely on iterative solver tolerances ($\mathcal{P}_h^\delta$) rather than exact projection. Second, generalizing across unseen observation operators (e.g., velocity magnitude or vorticity) still requires test-time refinement (APLR).

\bibliographystyle{unsrt}  
\bibliography{references}

\end{document}